\documentclass[sigconf]{acmart}

\usepackage[utf8]{inputenc} 
\usepackage[T1]{fontenc}    
\usepackage{hyperref}       
\usepackage{url}            
\usepackage{booktabs}       
\usepackage{amsfonts}       
\usepackage{nicefrac}       
\usepackage{microtype}      

\usepackage{wrapfig,lipsum,booktabs}
\usepackage{multirow}

\usepackage{hyperref}

\usepackage{math_commands}

\usepackage{hyperref}
\usepackage{url}
\usepackage{multirow}
\usepackage{graphicx}
\usepackage{booktabs}
\usepackage{caption}
\usepackage{graphicx}
\usepackage{subfigure}
\usepackage{placeins}

\usepackage{algorithm} 
\usepackage{algpseudocode}  

\usepackage{amsthm}

\usepackage{enumitem}
\usepackage{bm}
\newcommand{\eat}[1]{}

\newtheorem{problem}{Problem}
\theoremstyle{definition}

\usepackage{xspace}
\PassOptionsToPackage{usenames,dvipsnames}{xcolor}
\usepackage[usenames,dvipsnames]{xcolor}

\mathchardef\mhyphen="2D

\newcommand{\method}{Power-Link\xspace}

\newcommand{\xhdr}[1]{\noindent{{\bf #1.}}}

\usepackage{dsfont}

\def\rot{\rotatebox{72}}
\usepackage{pifont}
\newcommand{\xmark}{\ding{53}}%

\usepackage{ulem}
\usepackage{soul}
\definecolor{denim}{rgb}{0.173, 0.443, 0.729}
\newcommand{\revision}[1]{{{\textcolor{black}{#1}}}} 
 
\definecolor{chestnut}{cmyk}{0, 0.7808, 0.4429, 0.1412}

\definecolor{ForestGreen}{rgb}{0.133, 0.545, 0.133}
\definecolor{RoyalBlue}{rgb}{0.188, 0.15, 16.662}

\usepackage{adjustbox}
\usepackage{float}
\usepackage{booktabs}

\AtBeginDocument{%
  }

\copyrightyear{2024}
\acmYear{2024}
\setcopyright{acmlicensed}\acmConference[KDD '24]{Proceedings of the 30th ACM SIGKDD Conference on Knowledge Discovery and Data Mining}{August 25--29, 2024}{Barcelona, Spain}
\acmBooktitle{Proceedings of the 30th ACM SIGKDD Conference on Knowledge Discovery and Data Mining (KDD '24), August 25--29, 2024, Barcelona, Spain}
\acmDOI{10.1145/3637528.3671683}
\acmISBN{979-8-4007-0490-1/24/08}

\begin{document}

\title{Path-based Explanation for Knowledge Graph Completion}

\author{Heng Chang}
\authornote{The two first authors made equal contributions.}
\email{changh.heng@gmail.com}
\orcid{0000-0002-4978-8041}
\affiliation{%
  \institution{Huawei Technologies Co., Ltd.}
  \city{Beijing}
  \country{China}
}

\author{Jiangnan Ye}
\authornotemark[1]
\orcid{0009-0005-1936-5944}
\email{jiangnan.ye.aca@gmail.com}
\affiliation{%
  \institution{Huawei Technologies Co., Ltd.}
  \city{London}
  \country{United Kingdom}
}

\author{Alejo Lopez-Avila}
\orcid{0000-0002-4656-5109}
\email{alejo.lopez.avila@huawei.com}
\affiliation{%
  \institution{Huawei Technologies Co., Ltd.}
  \city{London}
  \country{United Kingdom}
}

\author{Jinhua Du}
\orcid{0000-0002-3267-4881}
\email{jinhua.du@gmail.com}
\affiliation{%
  \institution{Huawei Technologies Co., Ltd.}
  \city{London}
  \country{United Kingdom}
}

\author{Jia Li}
\orcid{0000-0002-6362-4385}
\authornote{Corresponding author.}
\email{jialee@ust.hk}
\affiliation{%
\institution{Hong Kong University of Science and Technology (Guangzhou)}
  \city{Guangzhou}
  \country{China}
}

\renewcommand{\shortauthors}{Heng Chang, Jiangnan Ye, Alejo Lopez-Avila, Jinhua Du, \& Jia Li}


\begin{abstract}
 Graph Neural Networks (GNNs) have achieved great success in Knowledge Graph Completion (KGC) by modelling how entities and relations interact in recent years. However, the explanation of the predicted facts has not caught the necessary attention. Proper explanations for the results of GNN-based KGC models increase model transparency and help researchers develop more reliable models. \eat{Considering the current practice for explaining KGC tasks all rely on instance/subgraph-based approaches, we argue that paths can provide more user-friendly and interpretable explanations of KGs.} \revision{Existing practices for explaining KGC tasks rely on instance/subgraph-based approaches, while in some scenarios, paths can provide more user-friendly and interpretable explanations. Nonetheless, the methods for generating path-based explanations for KGs have not been well-explored. To address this gap}, we propose Power-Link, the first path-based KGC explainer that explores GNN-based models. We design a novel simplified graph-powering technique, which enables the generation of path-based explanations with a fully parallelisable and memory-efficient training scheme.\eat{explanations with connection interpretability and sufficient path information.} We further introduce three new metrics for quantitative evaluation of the explanations, together with a qualitative human evaluation. Extensive experiments demonstrate that Power-Link outperforms the SOTA baselines in interpretability, efficiency, and scalability. The code is available at \url{https://github.com/OUTHIM/power-link}
\end{abstract}



\maketitle

\section{Introduction}\label{sec:intro}
Knowledge graphs (KGs) are a form of knowledge representation that encodes structured information about real-world entities and their relations as nodes and edges. 
Each edge connects two entities, forming a triple called a \textit{fact}.
However, real-world KGs often suffer from \textit{incompleteness}, which limits their usefulness and reliability. Incompleteness refers to not all possible facts being represented in the KG, either because they are unknown or unobserved. To address these issues, knowledge graph completion (KGC) is a task that aims to predict missing relations in a KG, given the existing facts~\cite{bordes2013translating,galarraga2015fast,ho2018rule}.

\begin{figure}[tb]
\centering
\includegraphics [width=\columnwidth]{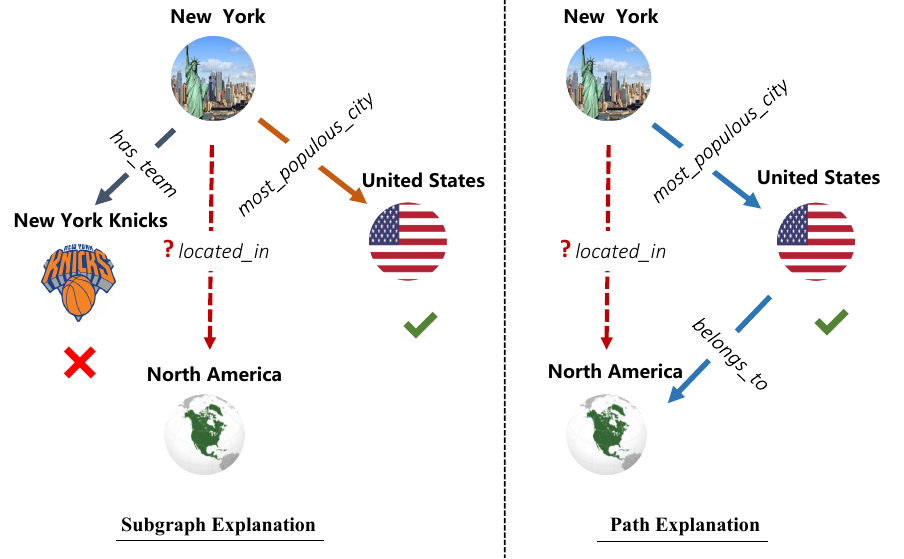}
\vspace{-6mm}
\caption{Example of the advantage that path-based explanation has over subgraph-based ones.}
\Description{Example of the advantage that path-based explanation has over subgraph-based ones.}
\label{fig:anchor-pic}
\end{figure}

GNN-based KGC models have shown effectiveness on the KGC task and attracted tremendous attention in recent years~\cite{ji2021survey}. While GNNs achieve remarkable success in completing KGs, how GNNs predict a given triplet candidate as factual remains unclear. As a result, the prediction needs substantial explanations before researchers and end users bring them into practice, which falls into the research scope of explainable artificial intelligence (XAI)~\cite{tjoa2020surveyXAI}. XAI enhances the transparency of black-box machine learning (ML) models and has been extensively investigated on GNNs for node/graph-level tasks on homogeneous graphs~\cite{taxonomy}. For a given data instance, these GNN explanation methods either learn a mask to select a subgraph induced by edges~\cite{pgexplainer,gnnexplainer} or search for a subgraph in the graph space~\cite{subgraphx} as an explanation.
However, to the best of our knowledge, the GNN explanation of KGs, especially regarding the KGC task, is surprisingly few in the literature.

\begin{figure*}[tb]
\centering
\includegraphics [width=0.9\textwidth]{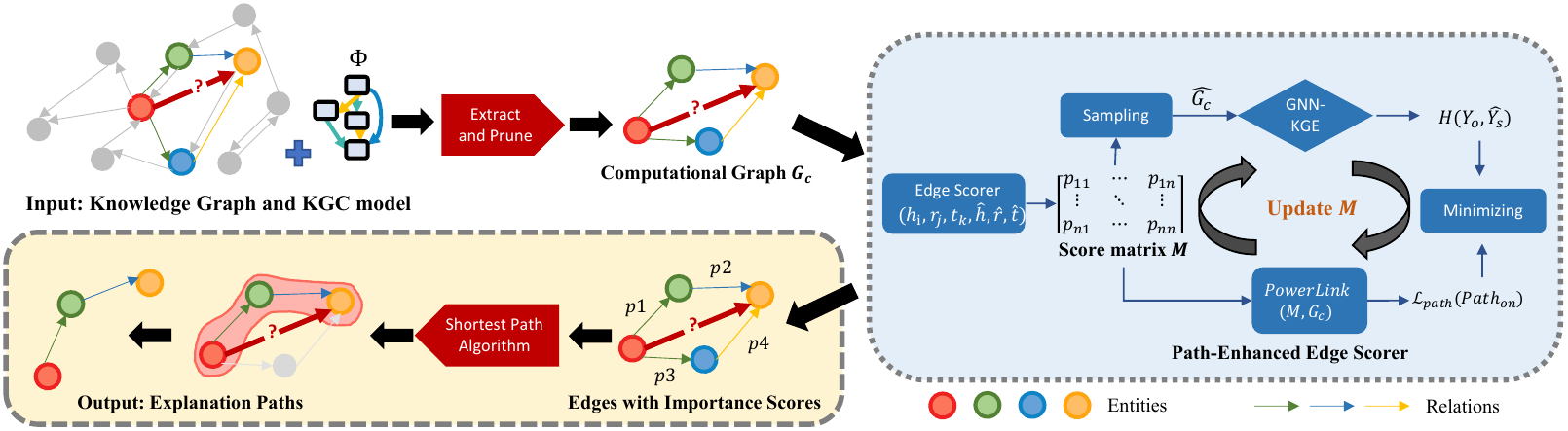}
\caption{The overall framework of the proposed \method. Given a KG and a trained KGC model as inputs, we aim to generate interpretable paths to explain why the KGC model predicts a target triplet is factual.}
\Description{The overall framework of the proposed \method. Given a KG and a trained KGC model as inputs, we aim to generate interpretable paths to explain why the KGC model predicts a target triplet is factual.}
\label{fig:framework}
\end{figure*}

Providing explanations for GNN-based KGC models is a novel and challenging task. Given the nature of KGs in modeling structured real-world information, their end users often have little or no background in AI or ML. Therefore, the explanations need to be more aligned with human intuitions. 
Existing approaches attempt to explain KGs from triplet or subgraph perspectives. Still, the problem of what constitutes a good explanation of KGs is more complex than on homogeneous graphs, especially in terms of satisfying the ``stakeholders’ desiderata''~\cite{langer2021we}. 
In contrast to the widely adopted instance/subgraph-based explanations, we focus on generating the explanations of GNN-based KGC models from the perspective of paths in this work:

\fbox{\begin{minipage}{25em}
\emph{Can paths in the KG provide better explanations for GNN-based embedding models on the KGC task?}
\end{minipage}}

Figure~\ref{fig:anchor-pic} illustrates three challenges of applying instance-based or subgraph-based GNN explanations from homogeneous graphs to KGs:
(1) \textit{Interpretability:} For KGs, explanations should reveal the connections between a pair of entities given a specific relation through paths on KGs. However, previous approaches produce disconnected subgraphs that do not show this connected pattern (e.g., left from Figure~\ref{fig:anchor-pic}). These subgraphs are not suitable for human interpretation and exploration.
(2) \textit{Sufficiency:} Paths contain sufficient information to explain whether a given triplet is factual. From Figure~\ref{fig:anchor-pic}, for explaining triplet $\langle New\_York, located\_in, North\_American \rangle$, the subgraph-based explanation brings in a noisy and irrelevant fact $\langle New\_York, has\_team, Knicks \rangle$. The instance-based one only extracts a related fact $\langle New\_York, most\_popular\_city, United\_States \rangle$ but loses sufficient information for users to interpret. Therefore, neither method is suitable for KGs where concise and informative semantics are preferred.
(3) \textit{Scalability:} Compared to general subgraphs, paths form a much smaller search space for optimal subgraph explanations since paths are simpler and scale linearly in the number of graph edges. Therefore, paths can better fit with KGs that usually come with massive entities.
To our best knowledge, there is only one approach, PaGE-Link~\cite{zhang2023page} that explores path-based solutions to the GNN explanation problem. But PaGE-Link focuses on heterogeneous graphs \revision{without considering the relation types and the scalability of the method, which makes it unsuitable to be directly applied to KGs.} 
Therefore, the question of KGs is still underexplored.

Given the important roles that paths play in the explanation of KGC tasks,
we propose \method to tackle the aforementioned challenges, which aims to \eat{close the gap between desired paths and explanations} \revision{enhance the interpretability, sufficiency, and scalability of the explanations by providing explainable paths} for the GNN-based KGC models. To the best of our knowledge, this is the first work specifically designed to explain KGC tasks via paths. Figure~\ref{fig:framework} depicts the overall framework of \method.
\revision{1) We first extract a k-hop subgraph around the source and target nodes and prune it to eliminate noisy neighbours.} 2) A Triplet Edge Scorer (TES) is proposed to measure the importance of each edge along the candidate paths \revision{by leveraging the node and edge type information}, which is specifically designed for the KG scenario. 
3) PaGE-Link employs the shortest-path searching algorithm \revision{to emphasize paths} in each iteration, which is computationally intensive and not applicable to large-scale KGs. It also introduces the accumulation of on-path(edges on the paths) errors during the learning process. To alleviate this issue, we propose to replace the shortest-path searching algorithm with a specially-designed graph-\textit{Power}-based method to amplify explainable on-path
edges. \revision{Our method is highly parallelisable with GPUs and consumes minimal memory resources, enabling easy scaling to large KGs.} The graph power-based approach also provides an interface for end users to select the desired path length of each explanation manually. \method generates interpretable paths to explain the prediction of GNN-based KGC models in an efficient manner. 
5) To better measure the performance of explanation methods on KGs from a quantitive perspective, we also propose three proper metrics, $Fidelity+$, $Fidelity-$, and $H\Delta R$, that reflect how good an explanation is in the empirical evaluation.

The main contributions of this paper are summarised as follows:
\begin{itemize}[noitemsep,topsep=0pt,parsep=0pt,partopsep=0pt]
    \item Motivated by the need to explain the prediction on the KGC task, we propose the first path-based methods \method for GNN-based KGC models that efficiently generate more human-interpretable explanations.
    \item We propose a novel path-enforcing learning scheme based on a simplified graph-powering technique, which enjoys high parallelisation and low memory cost. This enables \method to generate multiple explanatory paths precisely and efficiently.
    \eat{We present \method which reduces the explanation search space on KGs by a graph powering-based edge scorer, which excludes many less-meaningful subgraph candidates and generates multiple paths as explanations in parallel.}
    \item We propose three metrics to better evaluate the performance of path-based explanation methods.
    We show the superiority of \method in explaining GNN-based KGC models over SOTA baselines through extensive experiments from both quantitative and qualitative perspectives. 
\end{itemize}

\begin{table}[t]
\center
\caption{Methods and desired explanation properties of representative GNN and KGC explanation methods.}
\resizebox{\columnwidth}{!}{
\begin{tabular}{@{}lcccccc|c@{}}
\toprule
\multicolumn{1}{l}{Methods} & \rot{GNNExp~\scriptsize{\cite{gnnexplainer}}} & \rot{PGExp~\scriptsize{\cite{pgexplainer}}} & \rot{SubgraphX~\scriptsize{\cite{subgraphx}}} & \rot{PaGE-Link~\scriptsize{\cite{zhang2023page}}} & \rot{KE-X~\scriptsize{\cite{zhao2023ke}}} & \rot{Kelpie~\scriptsize{\cite{rossi2022explaining}}} & \rot{\method }\\ \midrule
Heterogeneity &\xmark &\xmark &\xmark &\checkmark &\checkmark &\checkmark &\checkmark \\
Explains GNN &\checkmark &\checkmark &\checkmark &\checkmark &\xmark &\xmark & \checkmark \\
Explains KGC &\xmark &\xmark &\xmark &\xmark &\checkmark &\checkmark &\checkmark \\
Path-based &\xmark &\xmark &\xmark &\checkmark &\xmark &\xmark &\checkmark \\
Scalability &\checkmark &\checkmark &\xmark &\checkmark &\xmark &\checkmark &\checkmark \\
Local Post-hoc &\xmark &\xmark &\xmark &\checkmark &\xmark &\checkmark &\checkmark \\
\bottomrule
\end{tabular}
}
\label{tab:related-work comp}
\Description{Power-Link wins on desired properties of good explanations.}
\end{table}

\section{Related Work}\label{sec:related-works}
\subsection{Knowledge Graph Completion (KGC)}
\eat{To shorten the paper, I deleted some parts of this section.}
Embedding-based KGC methods aim to embed the triplets into low-dimensional space such that triplets sharing similar patterns become closer to each other, \revision{which has achieved immense success in the past decade.} \eat{ Although the latter two categories are known for better transparency,} Please kindly refer to the survey~\cite{ji2021survey} for a thorough review of the recent advances in embedding-based methods.
Despite the success of GNN-based graph embedding models for KGC, the underlying mechanisms of how these embeddings facilitate KG completion are still unclear. Therefore, explaining KGC models is an important research challenge, especially under the post-hoc circumstance, which is the main focus of our paper. Specifically, we choose GNN-based KGC models as the targeting type of model to explain. \revision{GNNs are recently considered unnecessary for KGC due to their limited impact on the link prediction performance~\cite{zhang2022rethinking}. On the contrary, we present \method to underline the significant impact of GNNs on the explanation of KGC.}

\subsection{GNN and KGC Explanation}
\eat{Table~\ref{tab:related-work comp} compares the properties of \method with other representative methods for GNN and KGC Explanation. We highlight the main differences and advantages.} \revision{In this section, we review existing representative methods for GNN and KGC Explanation. For better clarity, we highlight their main differences and advantages by comparing them with our proposed \method in Table ~\ref{tab:related-work comp}.}

\xhdr{GNN explanation}
Explaining the prediction of GNNs~\cite{chang2020restricted,gu2020implicit,chang2021not,chang2021spectral,li2022semi,guan2021autogl} is an important task for understanding and verifying their behaviour. 
A common approach is to identify a subgraph that is most relevant to the prediction.
In this vein,
GNNExplainer~\cite{gnnexplainer} and PGExplainer~\cite{pgexplainer} use mutual information (MI) between the masked graph and the original prediction as the relevance measure and select edge-induced subgraphs by learning masks on graph edges and node features. SubgraphX~\cite{subgraphx} and GStarX~\cite{zhang2022gstarx} use game theory values, such as Shapley value~\cite{shapley} and structure-aware HN value~\cite{hn_value}, as the relevance measure, and select node-induced subgraphs by performing Monte Carlo Tree Search (MCTS) or greedy search on nodes. PaGE-Link~\cite{zhang2023page} argues that paths are more interpretable than subgraphs and extends the explanation task to the link prediction problem on heterogeneous graphs. 
For a review of other types of GNN explanations, please kindly check the surveys \cite{taxonomy,kakkad2023survey}.
Among them, our work is most closely related to PaGE-Link~\cite{zhang2023page}, as we also advocate for path-based explanations for GNNs. However, \revision{rather than the heterogeneous graphs,} we focus on a new challenging task of \revision{providing path-based explanations} for KGC. This could not be trivially tackled by the existing methods \revision{due to the lack of triplet information and the limitation of scalability}.

\xhdr{KGC explanation}
As the scope of this paper, the explanation of KGC is rather less explored especially compared with the fruitful results on GNN explanations.
~\citet{janik2022explaining} proposes ExamplE heuristics generate disconnected triplets as influential examples in latent space.
Similarly, \citet{zhao2023ke} leverages information entropy to quantify the importance of candidates by building a framework KE-X and generating explainable subgraphs accordingly.
~\citet{rossi2022explaining} develops an independent framework Kelpie to explain the prediction of embedding-based KGC methods by identifying the combinations of training facts.
However, these methods produce explanations in the form of subgraphs or triplets. \eat{what makes a good explanation is a complex topic.} We argue that paths provide better interpretability than unrestricted subgraphs/triplets. Thus path-based explanations are preferable on KGC tasks when the users have limited ML backgrounds.

\subsection{Paths on Graphs}
We briefly review the methods which leverage paths on graphs as crucial parts of their algorithms. Paths are often used to represent the structure and semantics of graphs, such as Katz index~\cite{katz1953new}, graph distance~\cite{bunke1998graph}, SimRank~\cite{jeh2002simrank}, and PathSim~\cite{sun2011pathsim}. They can also reveal the underlying relationship between a pair of nodes, and therefore bring more explanability to the methods. \cite{luo2021detecting} propose a context path-based GNN that recursively embeds the paths connecting nodes into the node embedding with attention mechanisms. In the same field, \citet{zhu2021neural} utilizes the idea from the neural bellman-ford algorithm to construct a general path-based framework NBFNet for both link prediction and KGC. \citet{zhu2022learning} further enhance the scalability of NBFNet by proposing to use A$*$ algorithm for path constructions. \citet{chang2023knowledge} propose to augment the knowledge used in the KGC task from the counterfactual view and enhance the explainability of the completion results. While these methods can generate non-trivial path-based explanations as a side-product, they cannot be applied to the black-box explanation setting in explainable artificial intelligence (XAI), where the pre-trained KG embeddings are given. Therefore, we argue that paths are still valuable for providing interpretable explanations for KGC under the local post-hoc scenario.

\section{Preliminaries}\label{sec:pre}
\subsection{Knowledge Graphs (KGs)}
A KG is a collection of triples of the form: 
\begin{equation}
    \notag \gG = (\gE, \gR) = \{(e_i,r_k,e_j)\} \\ \subset (\gE \times \gR \times \gE), 
\end{equation}
where $\gE$ denotes the set of entities (nodes) and $\gR$ is the set of relations (edges). $(e_i, e_j)$ are $i$-th and $j$-th entities and $r_k$ is $k$-th relation between them. The entities are indexed by $i$, $j$, and the relation is indexed by $k$. The number and type of relations in a KG can vary widely depending on the domain and the source of the data. We also use the notation $(h_i, r_k, t_j)$ to indicate an entity pair with a directional relation $r_k$, where $h_i$ is the head entity and $t_j$ is the tail entity.
\eat{Note that KGs can be viewed as a specialized version of heterogeneous graphs $\gG=(\gV, \gE, \phi, \tau)$ associated with a node type mapping function $\phi$ and an edge type mapping function $\tau$ can be viewed as the generalized type of KGs, where each node in KGs belongs to the same node type $\phi(v) = 1$ and each edge belongs to various of relation types $\tau(e) = |\gR|$. }

Throughout this paper, we use \textbf{bold} terms, $\mW$ or $\ve$, to denote matrix/vector representations for weights and entities, respectively. And we select \textit{italic} terms, $w_h$ or $\alpha$, to denote scalars.
We also use a binary adjacency matrix $\mA \in\{0,1\}^{|\gE|\times|\gE|}$ to define whether two entities are connected in a KG $\gG$. The $(i,j)$ entry $\mA_{ij}=1$ if $(h_i, t_j)$ is valid for any $r_k$ or otherwise $\mA_{ij}=0$.

\subsection{GNN-Based framework for KGC}\label{sec:pre_gnnkgc}
To predict the valid but unseen triplets, the task of KGC uses known triplets existing in the KG $\gG$.
GNN-Based framework dealing with KGC usually adopts an encoder-decoder framework~\cite{schlichtkrull2018modeling}, where GNNs perform as the encoder and the KGE score functions (e.g., TransE, DistMult, and ConvE) perform as the decoder. 
A KG's entities and relations are first embedded by the GNN encoder through the \textit{message passing} and \textit{neighbourhood aggregation} procedures~\cite{zhang2020deep}.
\eat{After having the embeddings from the GNN encoder for entities and relations, KGC models usually simulate how entities and relations interact with the help of a score function (decoder) $s:\gE\times\gR\times\gE\rightarrow \mathbb{R}$. }
Using the embeddings, the score function $s$, which is also referred to as the energy function in the energy-based learning framework, learns to assign a score $s(h_i,r_j,t_k)$ from either real space or complex space to each potential triplet $(h_i,r_j,t_k)\in\gE\times\gR\times\gE$. 
The learned scores reflect the possibility of the existence of triplets.

Because of the \textit{message passing} and \textit{neighborhood aggregation} mechanism in GNNs, a $L$-layer GNN only collects messages from the $L$-hop neighbors of an entity pair $e_i, e_j$ to compute their representation $\ve_i$ and $\ve_j$. Therefore, we constrain the search space of explaining the KGC model $\Phi(\gG, (h, r, t))$ with an $L$-Layer GNN encoder to the \textit{computation graph} $\gG_c = (\gE_c, \gR_c)$. $\gG_c$ is the $L$-hop ego-graph of the predicted pair $(s, t)$, i.e., the subgraph with entity set $\gE_c = \{e \in \gE | dist(e, e_i) \leq L \text{ or } dist(e, e_j) \leq L\}$. Therefore, the KGC result is thus fully determined by $\gG_c$, i.e., $\Phi(\gG, (h, r, t)) \equiv \Phi(\gG_c, (h, r, t))$. 

\section{Problem Formulation}
In this work, we focus on a \textit{post-hoc} and \textit{instance-level} KGC explanation problem. The post-hoc assumption means the model $\Phi(\cdot, \cdot)$ is already trained and fixed, and the explanation method does not modify its architecture or parameters. The instance-level assumption means that the explanation method generates an explanation for each individual prediction of a \textbf{target triplet}. We use the $\hat{hat}$ notation to denote the target to be explained and its related elements. The target triplet is denoted by $\targettriplet = (h_{\hat{i}}, r_{\hat{j}}, h_{\hat{k}})$. The explanation method aims to provide a rationale for why a given triplet is predicted as factual by the model $\Phi(\cdot, \cdot)$. In a practical KG user case, the explanation could address questions such as \emph{why a person’s nationality is USA}.
\eat{We narrow down our search space to the reduced computational graph $\gG_c$. We need further constraints to make the generated path concise and informative so as to align with natural human-interpretable explanations. Inspired by the definition from~\cite{zhang2023page}, we put two additional constraints on the path candidates: (1) The path should be short. Long paths are sub-optimal, as they may indicate overly complex connections that compromise the conciseness and persuasiveness of the explanation. Meanwhile, since GNN-based KGC models use neighbourhood aggregation to amplify the local similarity in the score function-induced embedding space, neighbours of an entity from multi-hops away have a much weaker influence on the prediction.  (2) The entities along the path should avoid high-degree nodes. Paths containing high-degree nodes are also less desirable because high-degree nodes are often generic, and a path going through them is not as informative. For example, the entity $Harry\_Potter\_and\_Half\_Blood\_Prince$ is connected to both $\langle *, belongs\_to, Film \rangle$ and $\langle *, genre, Fantasy \rangle$. When the KGC model predicts $Harry\_Potter\_and\_Half\_Blood\_Prince$ forms a fact with $\langle *, awarded, IGN\_Best\_Fantasy\_Film \rangle$, the path goes through the entity $Film$ provide mere information since $Film$ connects many irrelevant items. On the opposite, the entity $Fantasy$ with a much lower degree could form a better and more informative path explanation since it specifically reveals the property of the target entity.

Given the two assumptions and two constraints, } \revision{We narrow down the scope of explanation to paths. The explanatory paths should be concise and informative. It should only contain nodes and edges that are most influential to the prediction.} We formally define the explanation problem for GNN-based KGC models as:
\begin{problem}[Path-based KGC Explanation]
\label{prob:path}
We focus on the task of generating path-based explanations for a predicted fact between a pair of entities $e_i$ and $e_j$. Given a trained GNN-based KGC model $\Phi(\cdot, \cdot)$, a computation graph $\gG_c$ of $\hat{h},\hat{t}$ that extracted from the target KG $\gG$,
we find an explanation $\gP = $ \{ $\pi | \pi$ is a $\hat{h}-\hat{t}$ path with maximum length $L$ \}. \eat{and degree of each entity less than $D_{max}$ $|\gP|L \leq B$.} By proper construction of optimization on candidates from $\pi \in \gP$, we aim to generate concise and influential paths as explanations for the prediction.
\end{problem}

To further reduce the graph complexity and accelerate the process of finding paths, we adopt the \textit{$k$-core pruning} module from~\cite{zhang2023page} to eliminate spurious neighbours and improve speed.  The $k$-core of a graph is defined as the unique maximal subgraph with a minimum node degree $k$ \cite{kcore-def}, such that the $k$-core of $\gG_c$ is defined as $\gG_c^k = (\gE_c^k, \gR_c^k)$.

\section{Proposed Method: \method}\label{sec:method}
\method contains three components: a Triplet Edge Scorer (TES), a Path-Enforcing Learning module, and a Path Generation module. In a path-forcing way, the TES learns to create a score matrix for each edge in the graph. The path generation module then selects explanatory paths according to the score matrix. Next, we introduce them in detail.
\subsection{Triplet Edge Scorer}\label{sec:algorithm}
In the \method, we propose to learn a \textit{Triplet Edge Scorer (TES)} to leverage the entity and relation information in the KG. TES gives a score to every edge in the computational graph. The score measures the importance of each edge in explaining the prediction of the target triplet. It can also be interpreted as the probability of the existence of each edge for the explanation. Different from \cite{pgexplainer}, which only integrates node features to mark the edge scores, we consider the local meaning of each edge triplet to the explanation of the target triplet. To be more precise, we combine the triplet embedding vectors of the local edge $(h_i,r_j,t_k)$ with the embedding vectors of the explanation target $\targettriplet$. We use a Multi-Layer Perceptron(MLP) to process the combined features and generate the edge score. The overall formula of the TES can be defined as:
\begin{equation}
    TES(\cdot) = MLP\left(Combine(\cdot)\right)
\end{equation}
We propose two possible strategies for combining the local edge triplet and the target triplet:

\xhdr{Concatenation} In the first strategy, we simply concatenate the embedding vectors of both triplets. This strategy gives more flexibility to the scorer while resulting in more computational cost. \eat{Nonetheless, in the experiment, we found the running speed within the acceptable range, and this strategy performs slightly better.}The concatenation strategy is written as:
\begin{equation}
    Combine_{cat}(h_i, r_j, t_k,\hat{h}, \hat{r}, \hat{t}) = \left[ h_i \circ r_j \circ t_k \circ \hat{h} \circ \hat{r} \circ \hat{t} \right]
\end{equation}

\xhdr{Euclidean} In the second strategy, we manually calculate the Euclidean norm of the difference between the corresponding head, relation, and tail embeddings of the edge triplet and the target triplet. This results in a 3-dimensional vector. \revision{This strategy can be more beneficial to the KGC models with energy-based score functions.} \eat{We find this strategy works better for the TransE-based model, which implies that designing a strategy according to the score function of the GNN-based KGC model may render better performance. Even though, for the sake of generalization, we choose the concatenation strategy in our experiment.}The Euclidean strategy can be written as:
\begin{equation}
    Combine_{euc}(h_i, r_j, t_k,\hat{h}, \hat{r}, \hat{t}) = \left[\lVert h_i-\hat{h}\rVert, \lVert r_j-\hat{r}\rVert, \lVert t_k-\hat{t}\rVert \right]
\end{equation}

\subsection{Path-Enforcing Learning}
In order to train the edge scorer to identify explanatory edges, we propose a path-enforcing learning method relying on the simplified powering process of the score matrix. Our method is simpler, faster, more efficient, more customizable, and more stable compared with PaGE-Link\cite{zhang2023page}.  

We first briefly discuss the problem we find in the learning process of PaGE-Link. They optimize an explanatory weighted mask and enhance the path-forming explanations by simultaneously forcing the on-path~(edges \textbf{on} the paths between the target entities) weights to increase and the off-path~(edges \textbf{not} on the paths between the target entities) weights to decrease. The potential paths are selected using the shortest-path searching algorithm, whose cost matrix is designed based on the weighted mask and restrictions on the node degree. First, the shortest-path searching algorithm is not parallelisable, which is slow when scaled to large KGs. Second, we argue that this optimizing strategy may introduce noise at the early training stage. In the beginning, higher weights in the mask can be assigned to trivial edges because of incomplete training, while meaningful edges are ignored. Therefore, when applying shortest-path searching on the underfitting weighted mask, the algorithm may strengthen the meaningless paths and weaken the important explanatory paths. This introduces noise that can be propagated through epochs. We also observe that PaGE-Link often generates similar explanations as the GNNExplainer, which is not equipped with a path-enforcing training strategy. We consider this result from the early perturbations in the training process, which hinders the algorithm from finding explanatory paths.

To alleviate the early perturbations, we replace the shortest-path searching with a graph-power-based algorithm that enhances paths of specific lengths. The algorithm strengthens all on-path edges \eat{and debilitates all off-path edges}during the whole training process at a low cost of computational time. Different from the intuition of PaGE-Link, we find it unnecessary to suppress the off-path edges during training. This also gives us room to significantly improve the efficiency of the algorithm. By doing so, instead of powering the whole matrix with increasing sparsity, we only need to compute the parallelisable multiplication of a decreasing-sparsity row vector and a fixed-sparsity matrix. The model learns to balance the global explanatory performance and the forcing of paths from the initial stage of the training process without being affected by the early noise. Next, we explain the algorithm in detail.

\xhdr{Path Enforcing} 
Intuitionally, the $(i_{th}, k_{th})$ element in the probability adjacency matrix of power $l$ indicates the summation of the probabilities of all length-$l$ paths connecting nodes $(e_{i}, e_{k})$. Generally, we take the target element of the powered probability adjacency matrix and maximise it to strengthen all the edges on the path between target nodes. Next, we explain the idea in detail. Given an extracted and pruned computational graph $\gG_c^k = (\gE_c^k, \gR_c^k)$ with $|\nodeset|= N$, we let $\adj \in \mathbb{R}^{N \times N}$ be the adjacency matrix of the computational graph. Let $\mask = TES(r), r \in \gR_c^k$ be the probability matrix obtained by scoring all edges in the computational graph with TES. $\mask$ can be considered as a weighted adjacency matrix $\mask \in \mathbb{R}^{N \times N}$. Let $\mask_{\hat{i}\hat{k}}$ be the edge probability of the target triplet $\targettriplet$, indicating the value at the targeting position $(\hat{i}, \hat{k})$ of $\mask$. We write them as:
\begin{equation}
    \mask = 
    \begin{bmatrix}
        p_{11}       & \dots & p_{1N} \\
        \vdots      & \vdots & \vdots  \\
        p_{N1}       & \dots & p_{NN} 
    \end{bmatrix}, 
    \quad and \  p_{ik}=\mask_{ik}
\end{equation}
We suppose that we are interested in paths of length less than $L$. In order to enhance paths of length $l$ ($1 \leq l \leq L$), instead of powering the whole probability matrix, we use a \textbf{simplification trick}. We only power the target ($\hat{i}_{th}$) row vector in the matrix, which we call the \textbf{power vector} $\pvec = \mask_{\hat{i}:}, \ \pvec \in \mathbb{R}^{1 \times N}$. This is done by multiplying $\mask$ to $\pvec$ by $l-1$ times, yielding $\pvec^{(l)}$.\eat{To provide a clearer illustration, we break down a single time of this vector-matrix production.} 
We illustrate the product of the power vector and a single column in $\mask$ with Eq. (\ref{eq:power_hvec}). When multiplying $\pvec$ with the $k_{th}$ column of $\mask$,
\eat{we actually first multiply the probabilities of edges on each path between nodes $(\hat{h}, e_k)$. This gives the probability of every path connecting the node pair. We then sum up the path probabilities for each node pair and update the results back to the $k_{th}$ position on $\pvec$.}
we are actually updating the $k_{th}$ value in $\pvec$, $\pvec_{k}$, with the sum of the probabilities of all paths in length $l$, connecting nodes $(\hat{h}, t_k)$. 
\begin{equation}
    \pvec^{(l)}_{k} = \pvec^{(l-1)} \ \mask_{: k}
\label{eq:power_hvec}
\end{equation}
For better understanding, we present a simple example of $l=2, \hat{i}=3, k=4$:
\begin{equation}
    \pvec_{4}^{(2)} = p_{31}^{(1)}p_{14}^{(1)} + p_{32}^{(1)}p_{24}^{(1)} \dots + p_{3N}^{(1)}p_{N4}^{(1)}
\end{equation}
where $p$ indicates the probability of a single edge in the probability matrix $\mask$.
The general equation of the powering process for updating the whole power vector $\pvec$ for $l$ iterations can be written as:
\begin{equation}
    \pvec^{(l)} = \pvec^{(1)} \ \mask^{l-1}
\label{eq:general_power_hvec}
\end{equation}
We further normalize the power vector. We first divide it elementwisely by the number of paths between each node pair. This can be obtained by simply dividing the power vector by the $\hat{i}_{th}$ row of the adjacency matrix $\adj$ of power $l-1$. With the simplification trick, we only power the $\hat{i}_{th}$ row vector $\avec$. Similar to Eq. (\ref{eq:general_power_hvec}), we have $\avec^{(l)} = \avec^{(1)} \adj^{l-1}$. We then elementwisely take the root of it by the path length $l$. Now we have the average probability of the edge in the paths between the pair of nodes in each position of the matrix. The normalization process can be written as:
\begin{equation}
    \pvec^{(l)}_{k}=\sqrt[l]{\frac{\pvec^{(l)}_{k}}{\avec_{k}^{(l)}}}, \ 1 \leq k \leq |\gE| \
    \label{eq:avg_prob}
\end{equation}
We take the target $\hat{k}_{th}$ element $\pvec^{(l)}_{\hat{k}}$ from the powered power vector, corresponding to the connection to the tail node $\hat{t}$ in the target triplet. The element represents the average probability of the edges \textbf{on} the length $l$ paths between the target nodes $(\hat{i}, \hat{k})$. 

We iterate the above process. After each round $l$ of powering, we record the $\pvec^{(l)}_{\hat{k}}$. Finally, we get the average probability of all on-path edges of length less than or equal to $L$ by averaging all the $\pvec^{(l)}_{\hat{k}}$, denoted by $P_{on}$:
\begin{equation}\label{eq:on-off}
    P_{on} = \frac{1}{L-1}\sum_{l=1}^{L} \pvec^{(l)}_{\hat{k}}; 
\end{equation}
We design the path loss by maximizing the on-path probability, which is written as:
\begin{equation}
    \mathcal{L}_{path} = -\log (P_{on})
    \label{eq:path_loss}
\end{equation}

\xhdr{Mutual Information Maximising}
We use the same method in \cite{gnnexplainer} to guide the model to choose explanatorily important edges, which maximizes the mutual information between the predictions with selected edges and the predictions with the original edges. This is equivalent to minimizing the prediction loss, where $\odot$ denotes the elementwise product:
\begin{equation}
    \mathcal{L}_{prediction} = - \log P_{\Phi}\left(Y=1 \mid G = (\mask \odot \gG_{c}^k), \targettriplet \right)
    \label{eq:pred_loss}
\end{equation}

\xhdr{The Overall Loss}
We combine the path loss and prediction loss by summation. Besides, we also add a regularisation term on the mask (generated by TES), which is multiplied by a weight $\gamma$. The regularisation term plays a crucial role in further concentrating the TES on important edges. The overall loss is defined as:
\begin{equation}\label{eq:total}
    \mathcal{L}_{total} = \mathcal{L}_{prediction} + \mathcal{L}_{path} + \gamma \lVert \mask \rVert_2
\end{equation}

\subsection{Path Generation}
After we obtain the edge scorer $TES$ trained on the target triplet $\targettriplet$, we perform a final computation of the edge score matrix $\mask = TES(r), r \in \gR$. The score matrix contains the explanatory importance of every edge to the target triplet. We take the inverse values of the score matrix as the cost matrix and apply Dijkstra’s algorithm to obtain the shortest paths. The paths are the most important paths supporting the prediction of the target triplet.
\begin{equation}
    \pathset \leftarrow Dijkstra(\hat{h},\hat{t},\gG_{c}^k, \mask)
\end{equation}

Algorithm~\ref{alg:explain} describes the full process of \method. The learning process is highly parallelizable and can be accelerated with GPUs. Since the score matrix $\mask$ and the adjacency matrix are usually  sparse for KGs, by utilizing the sparse matrix operations, we are able to enjoy even faster and memory-efficient computation. We also provide a complexity analysis of \method and other methods in the Experiments section.

\begin{algorithm}
\caption{The overall algorithm for our proposed \method.}
\label{alg:explain}
\small
\begin{algorithmic}[1]
    \Require The KG $\graph$ with nodes $\nodeset$, edges $\edgeset$ and adjacency matrix $\adj$, GNN-based KGC model $\model$ trained on $\graph$, the parameteriser $TES$ that estimates the importance of each edge for the explanation, a target triplet $\targettriplet$ to be explained and the label $\hat{y}$, number of epochs $T$ for training the explanation mask, the maximum length $L$ of the explanation paths.
    \Ensure A set of paths $\pathset$ that explains the prediction.
    \State Extract the computation graph $\compg = \nodeset_{c}, \edgeset_{c}$;
    \State Prune the computation graph $\compg$ for the k-core $\compg^{k}$;
    \State $epoch = 1$, $P_{on}=0$;
    \While{$epoch <= T$}
        \State $l = 1$
        \State Calculate the score matrix $\mask = TES(r), r \in \gR_c^k$;
        \State Obtain the position $(\hat{i}, \hat{k})$ of the target triplet in 
 $\mask$
        \State Initialize the power vector $\pvec = \mask_{\hat{i}:}$
        \For{$l < L-1$}
            \State Compute the power of the power vector
            $  \pvec^{(l)} = \pvec^{(l-1)} \ \mask$;
            \State Compute the average edge probability by Eq.(\ref{eq:avg_prob});
            \State Accumulate the target on-path probability $P_{on} = P_{on} +  \pvec^{(l)}_{\hat{k}}$
        \EndFor
        \State Average the on-path probability $P_{on} = \frac{P_{on}}{L-1}$
        \State Compute the total loss by Eq.(\ref{eq:total});
        \State Update $TES$ through backpropagation;
    \EndWhile
    \State \Return $\pathset \leftarrow Dijkstra(\hat{h},\hat{t},\gG_{c}^k, \mask)$.
\end{algorithmic}
\end{algorithm}
\vspace{-3mm}

\section{Experiments}\label{sec:exps}
\subsection{Experimental Setup}
\xhdr{Datasets} We evaluate \method on the task KGC on the most popular datasets: FB15k-237~\cite{toutanova2015observed} and WN18RR~\cite{dettmers2018convolutional}.
Statistics of datasets can be referred to in the Appendix. We use the standard splits~\cite{toutanova2015observed, dettmers2018convolutional} for a fair comparison.

\xhdr{Baselines} We compare \method against three SOTA baselines. Two subgraph-based methods include GNNExplainer~\cite{gnnexplainer} and PGExplainer~\cite{pgexplainer}. One path-based method is PaGE-Link~\cite{zhang2023page}. Note that GNNExplainer and PGExplainer are not designed to provide path-based explanations. To accommodate them to our task, we modify them into GNNExplainer-Link and PGExplainer-Link by applying Dijkstra's Algorithm to their learned weighted masks to extract paths. We use these explanatory methods to explain the predictions of three popular types of GNN-based KGC models. The GNN encoders RGCN~\cite{schlichtkrull2018modeling},
WGCN~\cite{shang2019end} and CompGCN~\cite{vashishth2019composition} are respectively connected by the KGE decoders TransE, DistMult and ConvE. This gives 9 KGC models for each method to explain.

\begin{table*}[!h]
    \centering
    \caption{GNN-based KGC explanation results. $\uparrow$ indicates larger is better and $\downarrow$ means smaller is better. The best results are marked as \textbf{bold}. / indicates the metric is too insignificant to measure the explanation.}
    \label{tab:kg_result}
    \vspace{-4mm}
    \resizebox{0.95\textwidth}{!}{
        \begin{tabular}{llcccccc|cccccc}
            \toprule
             \multirow{2}{*}{\bf{Method}} & \multirow{2}{*}{\bf{Model}}
             & \multicolumn{6}{c}{\bf{FB15k-237}} & \multicolumn{6}{c}{\bf{WN18RR}} \\
             & & \bf{F+$\uparrow$} & \bf{F-$\downarrow$} & \bf{Sparsity$\uparrow$} & \bf{H$\Delta$R:1$\uparrow$} & \bf{H$\Delta$R:3$\uparrow$} & \bf{H$\Delta$R:5$\uparrow$} & \bf{F+$\uparrow$} & \bf{F-$\downarrow$} & \bf{Sparsity$\uparrow$} & \bf{H$\Delta$R:1$\uparrow$} & \bf{H$\Delta$R:3$\uparrow$} & \bf{H$\Delta$R:5$\uparrow$} \\
             \midrule
             
\multirow{9}{*}{\bf{GNNExp-Link}}
& RGCN + TransE      & 0.479           & 0.418          & 0.523           & 0.064          & 0.176          & 0.226          & 0.219          & 0.147          & 0.561          & \textbf{0.170} & 0.245 & \textbf{0.265} \\
& RGCN + DistMult    & 0.196           & 0.182          & 0.525           & 0.238          & 0.342          & 0.380          & 0.010          & 0.018          & 0.561          & 0.160          & 0.220 & \textbf{0.270} \\
& RGCN + ConvE       & 0.201           & 0.172          & 0.524           & 0.172          & 0.284          & 0.334          & 0.079          & 0.073          & 0.561          & \textbf{0.105} & \textbf{0.130} & \textbf{0.135} \\
& WGCN + TransE      & 0.019           & \textbf{0.001} & 0.562           & \textbf{0.378} & 0.506          & 0.592          & 0.124          & \textbf{0.001} & 0.556          & \textbf{0.675} & \textbf{0.710} & \textbf{0.715} \\
& WGCN + DistMult    & 0.003           & \textbf{0.002} & 0.562           & 0.558          & 0.696          & 0.772          & 0.001          & \textbf{0.001} & 0.562          & 0.125          & 0.290 & 0.345 \\
& WGCN + ConvE       & 0.001           & \textbf{0.001} & 0.562           & 0.376          & 0.596          & 0.682          & 0.004          & \textbf{0.001} & 0.562          & 0.180          & 0.365 & 0.450 \\
& CompGCN + TransE   & 0.002           & 0.003          & 0.562           & \textbf{0.014} & 0.034          & 0.044          & 0.003          & \textbf{0.003} & 0.563          & /              & /     & /     \\
& CompGCN + DistMult & 0.041           & 0.041          & 0.561           & \textbf{0.108} & 0.164          & 0.208          & 0.101          & 0.082          & 0.558          & /              & /     & /     \\
& CompGCN + ConvE    & 0.083           & 0.083          & 0.525           & 0.064          & 0.122          & 0.152          & 0.031          & \textbf{0.025} & 0.559          & /              & /     & /     \\

 \midrule
 \multirow{9}{*}{\bf{PGExp-Link}}
& RGCN + TransE      & 0.408           & 0.488          & 0.533           & 0.064          & \textbf{0.178} & 0.222          & 0.168          & 0.193          & 0.522          & \textbf{0.170} & 0.240 & \textbf{0.265} \\
& RGCN + DistMult    & 0.173           & 0.199          & 0.564           & \textbf{0.240} & 0.340          & 0.380          & 0.010          & \textbf{0.014} & 0.549          & 0.160          & 0.225 & 0.260 \\
& RGCN + ConvE       & 0.199           & 0.171          & 0.467           & 0.178          & 0.278          & 0.330          & 0.076          & 0.076          & 0.502          & \textbf{0.105} & 0.120 & 0.130 \\
& WGCN + TransE      & 0.756           & 0.760          & 0.526           & 0.346          & 0.538          & 0.614          & \textbf{0.189} & 0.188          & 0.479          & 0.540          & 0.650 & 0.675 \\
& WGCN + DistMult    & \textbf{0.686}  & 0.686          & 0.488           & 0.588          & 0.746          & 0.830          & \textbf{0.037} & 0.037          & 0.453          & \textbf{0.275} & \textbf{0.465} & 0.530 \\
& WGCN + ConvE       & \textbf{0.835}  & 0.836          & 0.548           & 0.450          & 0.678          & 0.744          & \textbf{0.327} & 0.326          & 0.529          & 0.295          & 0.460 & \textbf{0.530}\\
& CompGCN + TransE   & 0.002           & \textbf{0.002} & 0.487           & \textbf{0.014} & 0.038          & 0.046          & 0.002          & 0.004          & 0.508          & /              & /     & /     \\
& CompGCN + DistMult & 0.042           & 0.042          & 0.499           & 0.094          & 0.162          & 0.208          & 0.087          & 0.096          & 0.529          & /              & /     & /     \\
& CompGCN + ConvE    & 0.082           & 0.082          & 0.501           & 0.066          & 0.114          & 0.134          & 0.028          & 0.027          & 0.491          & /              & /     & /     \\ 
 
\midrule
\multirow{9}{*}{\bf{PaGE-Link}}
& RGCN + TransE      & 0.496           & 0.400          & \textbf{0.626}  & 0.066          & 0.170          & \textbf{0.236} & 0.190          & 0.179          & 0.737          & \textbf{0.170} & 0.240 & \textbf{0.265} \\
& RGCN + DistMult    & 0.212           & \textbf{0.178} & 0.653           & 0.220          & \textbf{0.364} & 0.430          & 0.010          & 0.018          & \textbf{0.713} & \textbf{0.165} & 0.225 & \textbf{0.270} \\
& RGCN + ConvE       & 0.232           & 0.175          & 0.662           & 0.180          & 0.308          & 0.378          & 0.008          & 0.088          & \textbf{0.713} & 0.100          & 0.125 & \textbf{0.135} \\
& WGCN + TransE      & 0.030           & 0.767          & 0.660           & 0.340          & 0.540          & 0.636          & 0.166          & 0.221          & 0.709          & 0.555          & 0.675 & 0.690 \\
& WGCN + DistMult    & \textbf{0.686}  & 0.683          & 0.644           & 0.588          & 0.746          & 0.832          & 0.036          & 0.037          & 0.719          & \textbf{0.275} & 0.455 & 0.525 \\
& WGCN + ConvE       & 0.491           & 0.490          & 0.664           & 0.374          & \textbf{0.687} & \textbf{0.788} & 0.325          & 0.327          & \textbf{0.715} & 0.290          & 0.460 & 0.525 \\
& CompGCN + TransE   & 0.001           & 0.004          & 0.638           & \textbf{0.014} & \textbf{0.040} & 0.046          & 0.001          & 0.004          & 0.721          & /              & /     & /     \\
& CompGCN + DistMult & 0.040           & 0.042          & \textbf{0.672}  & 0.092          & \textbf{0.176} & 0.218          & 0.046          & 0.116          & 0.714          & /              & /     & /     \\
& CompGCN + ConvE    & 0.096           & 0.074          & 0.649           & \textbf{0.068} & 0.118          & \textbf{0.164} & 0.030          & 0.028          & 0.718          & /              & /     & /     \\
 
\midrule
\multirow{9}{*}{\bf{\method}}
& RGCN +  TransE     & \textbf{0.699}  & \textbf{0.088} & 0.531           & \textbf{0.074} & 0.174          & 0.234          & \textbf{0.420} & \textbf{0.071} & \textbf{0.816} & \textbf{0.170} & \textbf{0.245} & 0.260 \\
& RGCN + DistMult    & \textbf{0.377}  & 0.186          & \textbf{0.784}  & 0.234          & 0.354          & \textbf{0.432} & \textbf{0.024} & 0.035          & 0.399          & \textbf{0.165} & \textbf{0.230} & 0.265 \\
& RGCN + ConvE       & \textbf{0.435}  & \textbf{0.111} & \textbf{0.832}  & \textbf{0.190} & \textbf{0.326} & \textbf{0.396} & \textbf{0.085} & \textbf{0.063} & 0.547          & \textbf{0.105} & \textbf{0.130} & 0.120 \\
& WGCN + TransE      & \textbf{0.775}  & 0.150          & \textbf{0.753}  & 0.376          & \textbf{0.582} & \textbf{0.648} & 0.120          & 0.232          & \textbf{0.938} & 0.550          & 0.655 & 0.680 \\
& WGCN + DistMult    & 0.646           & 0.294          & \textbf{0.672}  & \textbf{0.629} & \textbf{0.772} & \textbf{0.836} & 0.036          & 0.062          & \textbf{0.734} & \textbf{0.275} & 0.460 & \textbf{0.535} \\
& WGCN + ConvE       & 0.416           & 0.287          & \textbf{0.720}  & \textbf{0.474} & 0.646          & 0.762          & 0.324          & 0.184          & 0.625          & \textbf{0.300} & {\textbf0.470} & 0.525 \\
& CompGCN + TransE   & \textbf{0.0025} & 0.0026         & \textbf{0.7247} & \textbf{0.014} & \textbf{0.04}  & \textbf{0.056} & \textbf{0.004} & \textbf{0.003} & \textbf{0.872} & /              & /     & /     \\
& CompGCN + DistMult & \textbf{0.043}  & \textbf{0.025} & 0.506           & 0.092          & 0.170          & \textbf{0.220} & \textbf{0.135} & \textbf{0.015} & \textbf{0.763} & /              & /     & /     \\
& CompGCN + ConvE    & \textbf{0.137}  & \textbf{0.050} & \textbf{0.794}  & \textbf{0.068} & \textbf{0.130} & 0.154          & \textbf{0.041} & \textbf{0.025} & \textbf{0.942} & /              & /     & /     \\
            \bottomrule
        \end{tabular}
    }
\end{table*}

\begin{figure*}[tb]
\centering
\vspace{-2mm}
\includegraphics [width=0.9\textwidth]{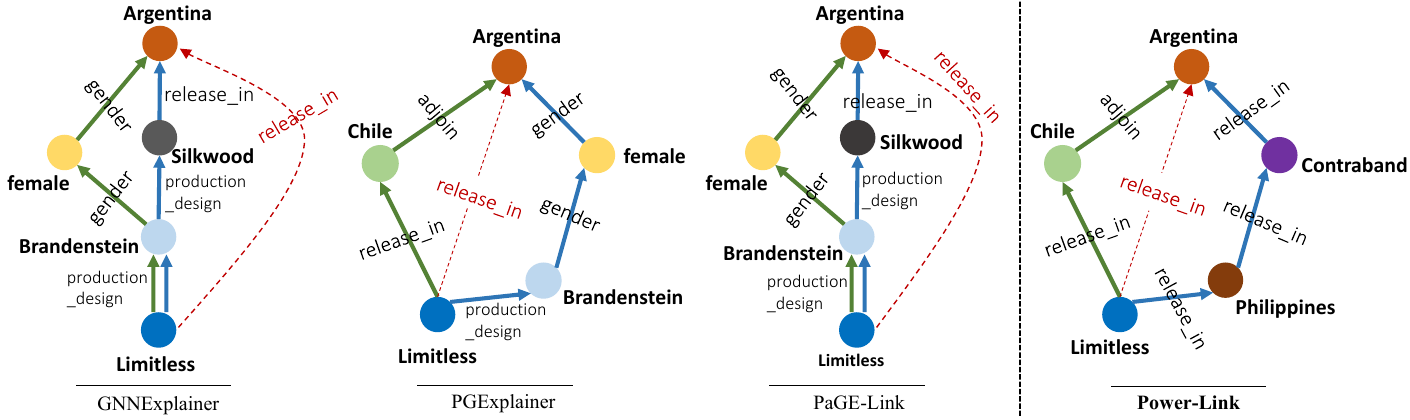}
\vspace{-4mm}
\caption{The explanations ({\textcolor{ForestGreen}{green}} and {\textcolor{RoyalBlue}{blue}} arrows) by different explainers for the prediction $\langle Limitless, release\_in, Argentina \rangle$ (\textcolor{red}{dashed red}). \method explains the fact by the $release\_in$ relationship, whereas baseline explanations are less interpretable.}
\Description{The explanations ({\textcolor{ForestGreen}{green}} and {\textcolor{RoyalBlue}{blue}} arrows) by different explainers for the prediction $\langle Limitless, release\_in, Argentina \rangle$ (\textcolor{red}{dashed red}). \method explains the fact by the $release\_in$ relationship, whereas baseline explanations are less interpretable.}
\label{fig:visualization}
\end{figure*}

\xhdr{Implementation Details} For the sake of better generalization, we choose the concatenation combination strategy in our experiment for TES. \eat{Our implementation is similar to~\cite {zhang2023page}, but we adapt it to fit the KG settings.}We follow the common setting of only explaining edges that the KGC model considers to exist. We call these edges \textit{explainable edges}.  For the FB15K237 dataset, we consider the edges that the KGC model assigns a score larger than 0.5 as explainable edges. We randomly sampled 500 explainable edges from the test set. The WN18RR dataset is much sparser and contains fewer samples in the test set. Just a few edges are scored higher than 0.5. Thus, we consider edges that have scores ranking at one as explainable edges and randomly sampled 200 target triplets. To ensure a fair comparison, target triplets to be explained from the same KGC model are identical for all explanation methods. We assume that longer paths contain less meaningful information for explanation and also increase the computational cost. Therefore, we choose a power order of 3 for the \method throughout the experiments, which yields an enhancement on the explanatory paths of length less than or equal to 3. A more detailed experiment setup can be found in the Appendix.

\subsection{Evaluation Metrics}
Motivated by \cite{amara2022graphframex}, we evaluate the learned explanation masks with $Fidelity+$, $Fidelity-$, and $Sparsity$. As the ranking difference is often considered important in KG-related tasks, we compare the quality of the explanation paths by calculating the times that the ranking of the target triplet drops after we delete the edges on the paths. This is denoted as $H\Delta R$. The detailed definitions of the metrics are introduced below:

\xhdr{Fidelity+ (F+)}
Given a target triplet $\targettriplet$ and the computational graph $\gG_c$, the explanatory subgraph $\gG_s$ is obtained by imposing the corresponding explanation mask on the computational graph, denoted by $\gG_s = \mask \cdot \gG_c$. Let $\hat{y}^\gG$ be the output triplet score of the GNN-based KGC model propagating on $\gG$, e.g. $\hat{y}^\gG = \Phi(\hat{h}, \hat{r}, \hat{t}, \gG)$. $Fidelity+$ measures the soft score difference between the prediction on the computational graph $\gG_c$ and the prediction on the computational graph excluding the explanatory subgraph $\gG_{c \backslash s}$. For $N$ test triplets, $Fidelity+$ can be written as:
\begin{equation}
    F_{+}=\frac{1}{N} \sum_{i=1}^N \left|\hat{y}^{\gG_c}_i - \hat{y}_i^{\gG_{c \backslash s}}\right|
\end{equation}
A good explanation should capture all the meaningful contributions to the prediction while ignoring the parts that are invaluable. Thus, a high $Fidelity+$ score can be an indication of a good explanation.

\xhdr{Fidelity- (F-)}
Under the same settings as $Fidelity+$, $Fidelity-$ measures the soft score difference of the prediction on the computational graph $\gG_c$ and the prediction on the explanatory subgraph $\gG_s$.For $N$ test triplets, $Fidelity-$ can be written as:
\begin{equation}
    F_{-}=\frac{1}{N} \sum_{i=1}^N \left|\hat{y}_i^{G_s} - \hat{y}^{\gG_c}_i\right|
\end{equation}
A good explanation should capture all the meaningful parts of the prediction. The prediction score on the explanatory graph should be close to that on the computational graph. Therefore, a low $Fidelity-$ score can be an indication of a good explanation.

\begin{figure}[htbp]
\centering
\includegraphics [width=0.95\columnwidth]{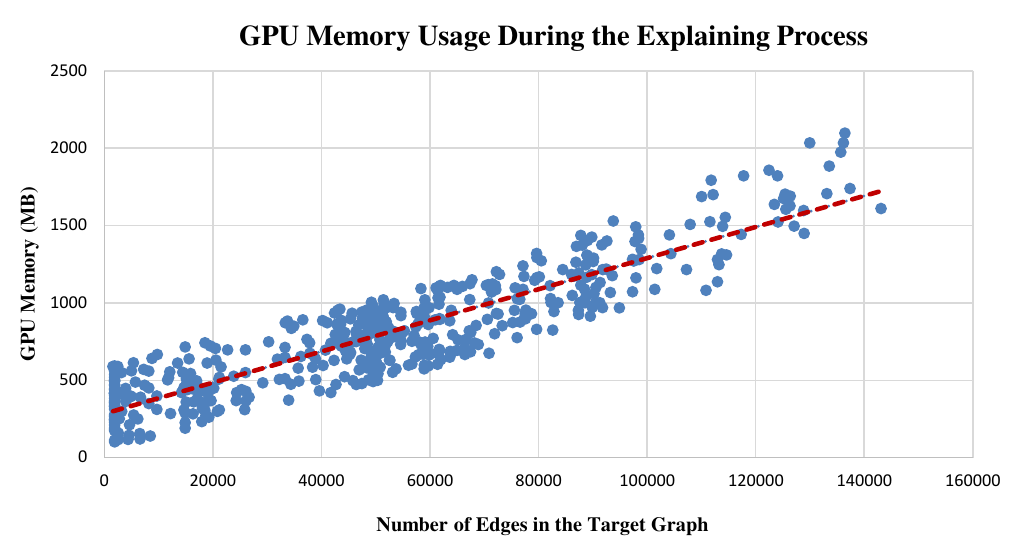}
\vspace{-4mm}
\caption{The GPU memory usage of \method during the explaining process against the number of edges in each graph. 500 samples are explained. For better illustration, we only record the memory usage of the explaining module. The memory occupied by the KGC model is ignored.}
\Description{The GPU memory usage of \method during the explaining process against the number of edges in each graph. 500 samples are explained. For better illustration, we only record the memory usage of the explaining module. The memory occupied by the KGC model is ignored.}
\label{fig:memory}
\end{figure}

\xhdr{Sparsity}
As the Fidelity score is often positively correlated with Sparsity, we consider Sparsity as one of our evaluation metrics. In our experiment, we measured the soft sparsity of the explanatory mask. For a good explanation, the explanatory mask should be highly condensed on the meaningful edges, therefore rendering high sparsity.

\xhdr{H$\Delta$R}
 Given a test triplet $(h, r, t)$, the computational graph $\gG_c$ and the explanatory path set $\pathset$, we extract a test graph $\gG_t$ by removing edges on the explanatory paths from the computational graph, denoted by $\gG_t = \{\nodeset_c, \edgeset_t\}, \ \edgeset_t \notin \pathset$. We let the GNN-based KGC model propagate on both the test graph and the computational graph. This gives two output scores of the same test triplet denoted by $\hat{y}^{G_c} = \Phi(h,r,t,\gG_c)$ and $\hat{y}^{G_t} = \Phi(h,r,t,\gG_t)$. $H \Delta R$ measures the hit rate when the ranking of $\hat{y}^{G_t}$ is smaller than $\hat{y}^{G_c}$. The hit indicates the ranking of the target triplet drops after we remove the explanatory paths. A good path-based explanation should include only important and meaningful paths and thus cause confidence drops to the model if we remove these paths, yielding a high hit rate. For $N$ test triplets, $H \Delta R$ can be written as:
 \begin{equation}
     H \Delta R (h,r,t,\pathset) = 1 - \frac{1}{N} \sum_{i=1}^N \mathds{1} \left(\hat{y}^{G_t}_i > \hat{y}^{G_c}_i\right)
 \end{equation}
 We aim to measure $H \Delta R$ against different numbers of explanatory paths removed from the computational graph. We denote $H \Delta R$ with $m$ paths removed as $H \Delta R:m$.

\section{Results and Analysis}
\subsection{KGC Explanation Results}
Table~\ref{tab:kg_result} summarises the explanation results of \method over three comparable baselines. We can have the following observations: (1) Our proposed \method achieves consistently better performance over all baselines on the two representative datasets. This indicates the effectiveness of the \method in generating good explanations for the KGC task. The superiority is more obvious on the FB15K237 KG, which is denser and more complicated. This reveals that the \method can be more effective at explaining complex KGs than sparse KGs. (2) Both path-based methods PaGE-Link and \method perform better than the two subgraph-based methods based on the path-oriented metrics. \revision{This aligns with the expected impact of the path-enforcing learning scheme.}\eat{This is consistent with our analysis of the advantages that paths have over subgraphs from Section~\ref{sec:intro}.} (3) The different choices of score functions exhibit similar trends in all methods. This indicates the stability of our \method regarding score function designs. (4) RGCN-based and WGCN-based models achieve higher scores in all our metrics than CompGCN-based ones. We assume that it is generally easier to explain the GNNs that strengthen the edge differences. This may be a future inspiration for designing explainability-reinforced GNNs for KGC. \eat{(5) Our proposed three metrics $Fidelity+$, $Fidelity-$, and $Sparsity$ could successfully reflect the evaluation of explanation on KGs given their results show similar trends as conventional $H\Delta R$ metrics.}(5) The explanation of WN18RR is harder than FB15k-237, as we can observe that the results on WN18RR are consistently lower. We assume that this is mainly because of the smaller and sparser graph of WN18RR. As a result, the meaningful paths are shorter and fewer, which makes it harder to form interpretable explanations.

\begin{table}[!t]
\centering
\caption{Runtime comparison between PaGE-Link and \method. We use the two methods to explain 500 samples predicted by WGCN-ConvE from FB15K237. The models are run on a single V100-32G GPU. Avg. indicates the average number per graph.}
\vspace{-3mm}
\label{tab:runtime}
\resizebox{\columnwidth}{!}{
\begin{tabular}{@{}llllll@{}}
\toprule
\textbf{Method} & \textbf{Avg. nodes} & \textbf{Avg. edges}  & \textbf{Avg. runtime} & \textbf{Total runtime} \\ \midrule
PaGE-Link       & \multirow{2}{*}{2335} & \multirow{2}{*}{51649} & 7.99 s                & 68m19s                 \\
Power-Link      &                       &                        & 2.369 s                & 21m28s                \\ \bottomrule         
\end{tabular}}
\end{table}

\begin{table}[!t]
\caption{Ablation study on the path-enforcing module. (-MI) indicates the model is explained with the mutual information loss only. The asterisk * indicates the results from Table \ref{tab:kg_result}.}
\label{tab:ab-path-enforcing}
\centering
\vspace{-3mm}
\resizebox{\columnwidth}{!}{
\begin{tabular}{@{}lllllll@{}}
\toprule
\textbf{Model}      & \textbf{F+}    & \textbf{F-}    & \textbf{Sparsity} & \textbf{H$\Delta$R:1} & \textbf{H$\Delta$R:3} & \textbf{H$\Delta$R:5} \\ \midrule
RGCN-DistMult-MI    & 0.356          & \textbf{0.168} & \textbf{0.834}    & 0.173          & 0.308          & 0.396          \\
WGCN-DistMult-MI    & 0.325          & 0.397          & 0.640             & 0.599          & 0.683          & 0.739          \\
CompGCN-DistMult-MI & 0.0411         & 0.0299         & \textbf{0.6556}   & 0.072          & 0.126          & 0.194          \\
*RGCN-DistMult      & \textbf{0.377} & 0.186          & 0.784             & \textbf{0.234} & \textbf{0.354} & \textbf{0.432} \\
*WGCN-DistMult      & \textbf{0.646} & \textbf{0.294} & \textbf{0.672}    & \textbf{0.629} & \textbf{0.772} & \textbf{0.836} \\
*CompGCN-DistMult   & \textbf{0.043} & \textbf{0.025} & 0.506             & \textbf{0.092} & \textbf{0.170} & \textbf{0.220} \\ \bottomrule
\end{tabular}
}
\end{table}

\subsection{Ablation Studies}
We conduct ablation studies using the KGC models trained on the FB15K-237 dataset to verify the impact of the design components in our proposed method. 

\xhdr{The path-enforcing module}
As shown in Table~\ref{tab:ab-path-enforcing}, we observed an obvious drop in path explanation performance when the path-enforcing module was removed. This is as expected since there is no strengthening on the impact of paths when explanations are generated without the path-enforcing. We expect the gap to be larger for denser graphs.

\xhdr{The combination method in TES}
Table~\ref{tab:ab-comb} shows that using the Euclidean combination method yields better explanation performance than Concatenation on RGCN-based KGC models. The superiority is more obvious for RGCN-TransE, which uses the energy-based loss function during KGC training. On the contrary, Euclidean downgrades the performance on WGCN-based KGC models. 

\xhdr{The mutual information loss}
We test the impact of mutual information loss(MIL) in explaining the WGCN-Distmult model. Table~\ref{tab:ab-ml} shows that without the mutual information loss, there is a significant drop in the explanation performance. The MIL is the key component that enables the explanation method to adjust the weighted mask of the KG according to the prediction of the KGC model. Without MIL, the path-enforcing module only strengthens the edges on the paths between the head and tail nodes without considering their impacts on the model's prediction. 

\xhdr{The order of powering}
Table~\ref{tab:ab-power-order} presents the explanation performance of Power-Link when we change the order of powering in the path-enforcing module. 
Interestingly, we find that increasing the powering order brings better performance and vice versa. We believe that increasing the powering order expands the range of length of the paths that the module strengthens. This can enable the generation of more explanatory paths that contain more important nodes. Even though, it is worth noting that longer paths may be less meaningful to the users, and higher powering order requires more computational power.

\xhdr{Comparison with PaGE-Link on heterogeneous graphs}
Since the path-enforcing learning module in \method can also be applied to heterogeneous graphs, for a better comparison, we reproduce the experiments in PaGE-Link with our proposed learning module. In both AugCitation and UserItemAttr datasets, \method achieves an improvement of 0.05 on the AUC score and is around 1.5 times faster. The result aligns with previous experiments, showing that \method is more precise and efficient. The specific statistics can be found in the Appendix Table~\ref{tab:hetero_result}.

\begin{table}[htbp]
\centering
\caption{Ablation study on the combination methods in TES. The asterisk * indicates the results from Table \ref{tab:kg_result}. }
\label{tab:ab-comb}
\vspace{-3mm}
\resizebox{\columnwidth}{!}{
\begin{tabular}{@{}lllllll@{}}
\toprule
\textbf{Model}          & \textbf{F+}    & \textbf{F-}    & \textbf{Sparsity} & \textbf{H$\Delta$R:1} & \textbf{H$\Delta$R:3} & \textbf{H$\Delta$R:5} \\
\midrule
RGCN-TransE-euc         & \textbf{0.711} & 0.204          & \textbf{0.906}    & \textbf{0.480} & \textbf{0.640} & \textbf{0.717} \\
RGCN-DistMult-euc       & 0.266          & \textbf{0.133} & 0.763             & \textbf{0.334} & \textbf{0.455} & \textbf{0.468} \\
RGCN-ConvE-euc          & 0.189          & 0.206          & \textbf{0.936}    & \textbf{0.444} & \textbf{0.542} & \textbf{0.590} \\
WGCN-TransE-euc         & 0.761          & 0.444          & 0.648             & \textbf{0.382} & 0.538          & \textbf{0.656} \\
WGCN-DistMult-euc       & 0.628          & 0.375          & 0.580             & 0.594          & 0.737          & 0.822          \\
WGCN-ConvE-euc          & 0.794          & 0.377          & 0.581             & 0.416          & \textbf{0.650} & 0.726          \\
*RGCN + TransE-concat   & 0.699          & \textbf{0.088} & 0.531             & 0.074          & 0.174          & 0.234          \\
*RGCN + DistMult-concat & \textbf{0.377} & 0.186          & \textbf{0.784}    & 0.234          & 0.354          & 0.432          \\
*RGCN + ConvE-concat    & \textbf{0.435} & \textbf{0.111} & 0.832             & 0.190          & 0.326          & 0.396          \\
*WGCN + TransE-concat   & \textbf{0.775} & \textbf{0.150} & \textbf{0.753}    & 0.376          & \textbf{0.582} & 0.648          \\
*WGCN + DistMult-concat & 0.646          & \textbf{0.294} & \textbf{0.672}    & \textbf{0.629} & \textbf{0.772} & \textbf{0.836} \\
*WGCN + ConvE-concat    & 0.416          & \textbf{0.287} & \textbf{0.720}    & \textbf{0.474} & 0.646          & \textbf{0.762} \\
\bottomrule
\end{tabular}
}
\end{table}

\begin{table}[htbp]
\caption{Ablation study on the mutual information loss. (-PS) indicates that only path-searching loss is used during the explanation(without mutual information loss). The asterisk * indicates the results from Table \ref{tab:kg_result}. }
\label{tab:ab-ml}
\centering
\vspace{-3mm}
\resizebox{\columnwidth}{!}{
\begin{tabular}{@{}lllllll@{}}
\toprule
\textbf{Model}   & \textbf{F+}    & \textbf{F-}    & \textbf{Sparsity} & \textbf{H$\Delta$R:1} & \textbf{H$\Delta$R:3} & \textbf{H$\Delta$R:5} \\ \midrule
WGCN-DistMult-PS & 0.013          & 0.021          & 0.938             & 0.168          & 0.220          & 0.238          \\
*WGCN-DistMult   & \textbf{0.646} & \textbf{0.294} & \textbf{0.672}    & \textbf{0.629} & \textbf{0.772} & \textbf{0.836} \\ \bottomrule
\end{tabular}
}
\end{table}

\begin{table}[htbp]
\centering
\caption{Ablation study on the powering order. The asterisk * indicates the results from Table \ref{tab:kg_result}.}
\label{tab:ab-power-order}
\vspace{-3mm}
\resizebox{\columnwidth}{!}{
\begin{tabular}{@{}lrllllll@{}}
\toprule
\textbf{Model} & \multicolumn{1}{l}{\textbf{Order}} & \textbf{F+}    & \textbf{F-}    & \textbf{Sparsity} & \textbf{H$\Delta$R:1} & \textbf{H$\Delta$R:3} & \textbf{H$\Delta$R:5} \\ \midrule
RGCN-DistMult  & 4                                  & 0.367          & \textbf{0.172} & 0.803             & 0.224    & \textbf{0.376} & \textbf{0.440} \\
RGCN-DistMult  & 2                                  & 0.172          & 0.175          & \textbf{0.954}    & 0.200          & 0.296          & 0.350          \\
*RGCN-DistMult & 3                                  & \textbf{0.377} & 0.186          & 0.784             & \textbf{0.234}    & {0.354}    & {0.432}    \\ \bottomrule
\end{tabular}
}
\end{table}

\subsection{Visualization and Performance Analysis}
\eat{Add the running time + GPU memory experiment (one table and one figure)}
\eat {Move the table of Pagelink dataset to the appendix}
\xhdr{Visualization of explanations}
Figure~\ref{fig:visualization} depicts the explanations for the predicted fact $\langle Limitless, release\_in, Argentina \rangle$ by different explainers. We can find that only \method explains why the KGC result is factual by the most reasonable relationship $release\_in$ along both of the generated paths. All other baseline explainer includes noisy edges (\ie, $Brandenstein, gender, female$) that are rather irrelevant to the fact. We also find that PaGE-Link generates the same explanation paths as the GNNExplainer though the latter is not path-enforced. We attribute this to the noise introduced at the early training stage of PaGE-Link.

\xhdr{Runtime comparison}
We compare the runtime of \method and PaGE-Link in Table \ref{tab:runtime}. \method is 3.18 times faster in total runtime and 3.37 times faster in average runtime per graph than PaGE-Link. This shows that the parallelisability of \method significantly enhances the speed of the explaining process.

\xhdr{GPU memory usage}
In Figure \ref{fig:memory}, we illustrate the GPU memory usage during the explaining process against the number of edges. We can observe that the memory usage(in MB) is linearly related to the edge number, with a slope of around 0.01. When explaining a graph of up to 140k edges, \method only takes up around 2GB GPU memory. This proves the excellent scalability of \method.

\xhdr{Human evaluation}
We conduct a human evaluation to test the ability of the explaining methods to improve the transparency and interoperability of the KGC models. We randomly selected 100 samples from the predictions of the WGCN-ConvE model on FB15K237. We design a user interface with a layout similar to Figure \ref{fig:visualization}. The participants include AI researchers in NLP and recommendation systems with limited knowledge of KG and GNNs. They are asked to select the best explanation among the given 4, where multiple choices are allowed. 300 evaluations are collected, among which \textbf{64.7\%} consider \method as the best, 55.0\% support PaGE-Link, 58.7\% support PGExp-Link and 56.7\% support GNNExp-Link. The human evaluation further shows the superiority of \method over other baseline methods.

\eat{\xhdr{Ablation study on the path loss} We investigate four invariants of the path loss in Power-Link and compare them against the original design (baseline). The results are shown in Table~\ref{tab:ablation}. The observations are: (1) On-path loss only:
In this variant, we only maximize the $P_{on}$ during training. The loss becomes $\mathcal{L}_{path} = -\log (P_{on})$. We observe from the results that even though there is a slight increase on the $H \Delta R$, the Sparsity drops significantly along with the F+. This indicates a higher probability for the shortest-path searching algorithm to select noisy explanatory paths, which harms the robustness of the algorithm. (2) Off-path loss only:
In this variant, we only minimize the $P_{off}$ during training. We observe that scores of all metrics heavily decrease except for the sparsity. Thus, we conclude that the algorithm can hardly select meaningful paths by merely minimizing $P_{off}$.
(3) Explicit way of enforcing:
In this variant, we handle $P_{on}$ and $P_{off}$ in the same explicit way as PaGE-Link does $\mathcal{L}_{path} = - P_{on} + P_{off}$ We find an obvious drop over scores of all the metrics. This shows the advantage of our method over PaGE-Link in dealing with $P_{on}$ and $P_{off}$. 
(4) Path of length 2:
In this variant, we only power the mask $\mask$ for one time, which just enhances paths of length 2. We can see a prominent decline in the overall performance. This is as expected since paths of length three are ignored, which may contain valuable support for the prediction. Nonetheless, we suggest from the results that in some situations where computational cost matters, paths of length two can be sufficient to provide acceptable explanations.}

\section{Conclusion}\label{sec:conclusion}
In this paper, we propose \method that generates explanatory paths via a novel simplified graph-powering technique. Based on GNNs, \method is the first path-based explainer for KGC tasks. Our approach sheds light on the direction of model transparency on widely used KGs, especially given the practical importance of KGs in industrial deployment. Extensive experiments demonstrate both quantitatively and qualitatively that \method outperforms SOTA explainers in terms of interpretability, efficiency, and scalability. We hope our work can inspire future research to design better GNN-based frameworks that enhance the explainability of KGC models.

\begin{acks}
This work was partially supported by NSFC Grant No. 62206067 and Guangzhou-HKUST(GZ) Joint Funding Scheme 2023A03J0673.
\end{acks}

\bibliographystyle{ACM-Reference-Format}
\balance
\bibliography{refs}

\newpage
\appendix


\section{Computational complexity}
In Table~\ref{tab:complexity}, we summarize the time complexity of \method and representative existing methods for explaining a prediction with computation graph $\gG_c = (\gE_c, \gR_c)$ on a full graph $\gG = (\gE, \gR)$. Let $T$ be the mask learning epochs. GNNExplainer has complexity $|\gR_c|T$ as it learns a mask on $\gR_c$. PGExplainer has a training stage that covers edges in the entire graph and thus scales in $O(|\gR| T)$. KE-X adopts the subgraph-based approach for explanation and has similar time complexity as SubgraphX~\cite{subgraphx}, which is $\Theta(|\gE_c|\hat D^{2B - 2})$.
$\hat D$ is the maximum degree in $\gG_c$ and $B$ is a manually chosen budget for nodes.
For PaGE-Link, its complexity consists of two parts: linear complexity in $|\gR_c|$ for the k-core pruning step and $|\gR_c^k||\gE_c^k|T + |\gR_c^k|T$ for the mask learning with Dijkstra's algorithm and sparse matrix multiplication step. For our \method, the graph powering step involves a vector and a sparse matrix multiplication for $L$ times for a $L$-length path, therefore, has complexity $|\gR_c^k|L$. The k-core pruning has the same complexity, but we only need to do the shortest path searching once. Therefore, the total time complexity of \method is $O(|\gR_c| + |\gR_c^k|LT)$.
Therefore, \method has better time complexity over PaGE-Link and both are better than existing methods since $|\gR_c^k|$ is usually smaller than $|\gR_c|$. 
Both PaGE-Link and \method could converge faster, i.e., has a smaller $T$, due to the smaller space of candidate explanations from paths.

\begin{table}[htbp]
\center
\caption{Time complexity of \method and other methods.}\label{tab:complexity}
\resizebox{\columnwidth}{!}{
\begin{tabular}{@{}cccc|c@{}}
\toprule
GNNExp~\cite{gnnexplainer} & PGExp~\cite{pgexplainer} & KE-X~\cite{zhao2023ke} & PaGE-Link~\cite{zhang2023page} & \method  \\ \midrule    
$O(|\gR_c| T)$ & $O(|\gR| T)$ & $\Theta(|\gE_c|\hat D^{2B - 2})$ & $O(|\gR_c| + |\gR_c^k||\gE_c^k|T  + |\gR_c^k|T)$ & $O(|\gR_c| + |\gR_c^k|LT)$ \\
\bottomrule
\end{tabular}
}
\end{table}

\section{Notations}
Table~\ref{tab:notation} summarizes the notations that are used throughout the paper.

\begin{table}[htbp]
\center
\caption{Notation of main notations with description.}\label{tab:notation}
\resizebox{\columnwidth}{!}{
\begin{tabular}{@{}l|l@{}}
\toprule
\multicolumn{1}{l}{Notation} & Definition and description \\ \midrule
$\gG = (\gE, \gR)$ & a KG $\gG$, entity set $\gE$, and relation set
$\gR$ \\
$\mask$ & The explanatory mask contains scores for each edge in the graph. \\
$(h, r, t)$ &  a (head, relation, tail) triplet \\
$\mA$ & the adjacency matrix of KG $\gG$\\
$\gP$ & the set of paths\\
$\gP_{ij}^{L}$ & the set of paths of length L connecting a pair of nodes $(i,j)$\\
$\Phi(\cdot, \cdot)$ & the GNN-based KGC model to explain \\
$s:\gE\times\gR\times\gE\rightarrow \mathbb{R}$ &  the score function used for embedding learning \\
$\targettriplet$ & the triplet that is predicted to be fact by the KGC model \\
$\pvec$ & the power vector which is the $\hat{i}_{th}$ row vector in $\mask$ \\
$\avec$ & the adjacency power vector which is the $\hat{i}_{th}$ row vector in $\adj$ \\
$\ve, \vr$ & the representation of entity and relation \\
$Y = \Phi(\gG, \targettriplet)$ & the KGC prediction of the target triplet $\targettriplet$ \\
$\gG_c = (\gE_c, \gR_c)$ & the computation graph, i.e., L-hop ego-graph of $\targettriplet$ \\
$\gG_{c\backslash s} = (\gE_{c\backslash s}, \gR_{c\backslash s})$ & the computational graph but excluding the subgraph \\
\bottomrule
\end{tabular}
}
\end{table}

\section{Statistics of datasets}
Dataset statistics of FB15k-237 and WN18RR for KGC are summarized in Table~\ref{tab:kg_statistics}.

\begin{table}[b]
    \centering
    \caption{Dataset statistics for knowledge graph completion.}
    \label{tab:kg_statistics}
    \resizebox{\columnwidth}{!}{%
    \begin{tabular}{lccccc}
    \toprule
    \multirow{2}{*}{\bf{Dataset}} & \multirow{2}{*}{\bf{\#Entity}} & \multirow{2}{*}{\bf{\#Relation}} & \multicolumn{3}{c}{\bf{\#Triplet}} \\
    & & & \bf{\#Train} & \bf{\#Validation} & \bf{\#Test} \\
    \midrule
    FB15k-237~\cite{toutanova2015observed} & 14,541 & 237 & 272,115 & 17,535 & 20,466 \\
    WN18RR~\cite{dettmers2018convolutional} & 40,943 & 11 & 86,835 & 3,034 & 3,134 \\
    \bottomrule
    \end{tabular}
    }
\end{table}

\section{Metrics of the KGC models}
We provide the metrics of the KGC models used for explanation in Table~\ref{tab:kgc_metrics}.

\begin{table}[htbp]
\centering
\caption{Metrics of the KGC models used for explanation.}
\label{tab:kgc_metrics}
\vspace{-3mm}
\resizebox{\columnwidth}{!}{
\begin{tabular}{@{}clrrrr@{}}
\toprule
\multicolumn{1}{l}{\textbf{Dataset}} & \textbf{Model}   & \multicolumn{1}{l}{\textbf{MRR}} & \multicolumn{1}{l}{\textbf{Hit@1}} & \multicolumn{1}{l}{\textbf{Hit@3}} & \multicolumn{1}{l}{\textbf{Hit@10}} \\
\midrule
\multirow{9}{*}{FB15K-237}           & CompGCN-ConvE    & 0.352                            & 0.261                              & 0.385                              & 0.536                               \\
                                     & CompGCN-DistMult & 0.343                            & 0.252                              & 0.376                              & 0.524                               \\
                                     & CompGCN-TransE   & 0.230                            & 0.246                              & 0.368                              & 0.510                               \\
                                     & RGCN-ConvE       & 0.430                            & 0.376                              & 0.461                              & 0.524                               \\
                                     & RGCN-DistMult    & 0.380                            & 0.327                              & 0.414                              & 0.471                               \\
                                     & RGCN-TransE      & 0.210                            & 0.132                              & 0.228                              & 0.474                               \\
                                     & WGCN-ConvE       & 0.319                            & 0.235                              & 0.349                              & 0.487                               \\
                                     & WGCN-DistMult    & 0.272                            & 0.200                              & 0.295                              & 0.415                               \\
                                     & WGCN-TransE      & 0.288                            & 0.213                              & 0.315                              & 0.434                               \\
\midrule
\multirow{9}{*}{WN18RR}              & CompGCN-ConvE    & 0.431                            & 0.395                              & 0.439                              & 0.515                               \\
                                     & CompGCN-DistMult & 0.431                            & 0.395                              & 0.439                              & 0.515                               \\
                                     & CompGCN-TransE   & 0.206                            & 0.126                              & 0.276                              & 0.507                               \\
                                     & RGCN-ConvE       & 0.464                            & 0.389                              & 0.472                              & 0.533                               \\
                                     & RGCN-DistMult    & 0.331                            & 0.311                              & 0.348                              & 0.402                               \\
                                     & RGCN-TransE      & 0.182                            & 0.132                              & 0.203                              & 0.396                               \\
                                     & WGCN-ConvE       & 0.449                            & 0.413                              & 0.467                              & 0.521                               \\
                                     & WGCN-DistMult    & 0.320                            & 0.283                              & 0.337                              & 0.383                               \\
                                     & WGCN-TransE      & 0.118                            & 0.032                              & 0.165                              & 0.284                              \\
\bottomrule
\end{tabular}
}
\end{table}

\section{Reproduce PaGE-Link experiments with Power-Link}
\begin{table}[htbp]
\centering
\caption{AUC and Running time comparison between \method and PaGE-Link on AugCitation and UserItemAttr datasets. We report the average time over $5$ splits for each method here. $s$ denotes seconds and $m$ denotes minutes.}
\vspace{-4mm}
\label{tab:hetero_result}
\resizebox{\columnwidth}{!}{
\begin{tabular}{llccccc}
\toprule
\multirow{2}{*}{\bf{Method}} & \multicolumn{2}{c}{\bf{AugCitation}} & \multicolumn{2}{c}{\bf{UserItemAttr}} \\
& \bf{AUC}    & \bf{Running Time}  & \bf{AUC}  & \bf{Running Time} \\ 
\hline
PaGE-Link  & 0.966      & 33m19s           & 0.956       & 4m33s            \\
Power-Link & \textbf{0.971}      & \textbf{22m42s}           & \textbf{0.961}       & \textbf{3m30s} \\ \bottomrule
\end{tabular}
}
\end{table}

We reproduce the experiments in PaGE-Link with the path-enforcing learning module of \method. As a complement to the analysis presented in the Ablation Study section, the results of the experiments are listed in Table \ref{tab:hetero_result}.

\section{Experiment Setup}
We provide the detailed setup of the experiment in Table \ref{tab:exp_setup}.

\begin{table*}[htbp]
\centering
\caption{The detailed setup of the experiment}
\label{tab:exp_setup}
\vspace{-4mm}
\resizebox{\textwidth}{!}{
\begin{tabular}{@{}lllllllllll@{}}
\toprule
\textbf{Encoder} & \textbf{Decoder} & \textbf{Dataset} & \textbf{k-Hop} & \textbf{k-Core} & \textbf{Epoch} & \textbf{Learning Rate} & \textbf{Max Node Num} & \textbf{Sample Num} & \textbf{Regularisation Weight} & \textbf{Threshold}                 \\ \midrule
CompGCN                                       & TransE           & FB15K237         & 1              & 2               & 50             & 0.005                  & 1000                  & 500                 & 0.02                           & \textgreater 0.5   \\
CompGCN                                       & DistMult         & FB15K237         & 1              & 2               & 50             & 0.005                  & 1000                  & 500                 & 0.03                           & \textgreater 0.5   \\
CompGCN                                       & ConvE            & FB15K237         & 1              & 2               & 50             & 0.005                  & 1000                  & 500                 & 0.001                          & \textgreater 0.5   \\
RGCN                                          & TransE           & FB15K237         & 1              & 2               & 50             & 0.005                  & 1000                  & 500                 & 0.03                           & \textgreater 0.5   \\
RGCN                                          & DistMult         & FB15K237         & 1              & 2               & 50             & 0.005                  & 1000                  & 500                 & 0.03                           & \textgreater 0.5   \\
RGCN                                          & ConvE            & FB15K237         & 1              & 2               & 50             & 0.005                  & 1000                  & 500                 & 0.03                           & \textgreater 0.5   \\
WGCN                                          & TransE           & FB15K237         & 2              & 2               & 50             & 0.005                  & 5000                  & 500                 & 0.03                           & \textgreater 0.5   \\
WGCN                                          & DistMult         & FB15K237         & 2              & 2               & 50             & 0.005                  & 5000                  & 500                 & 0.03                           & \textgreater 0.5   \\
WGCN                                          & ConvE            & FB15K237         & 2              & 2               & 50             & 0.005                  & 5000                  & 500                 & 0.03                           & \textgreater 0.5   \\ \midrule
CompGCN                                       & TransE           & WN18RR           & 3              & 2               & 50             & 0.005                  & 2000                  & 200                 & 0.03                           & hit@1              \\
CompGCN                                       & DistMult         & WN18RR           & 3              & 2               & 50             & 0.005                  & 2000                  & 200                 & 0.1                            & hit@1              \\
CompGCN                                       & ConvE            & WN18RR           & 3              & 2               & 50             & 0.005                  & 2000                  & 200                 & 0.1                            & hit@1              \\
RGCN                                          & TransE           & WN18RR           & 3              & 2               & 50             & 0.005                  & 2000                  & 200                 & 0.03                           & hit@1              \\
RGCN                                          & DistMult         & WN18RR           & 3              & 2               & 50             & 0.005                  & 2000                  & 200                 & 0.04                           & hit@1              \\
RGCN                                          & ConvE            & WN18RR           & 3              & 2               & 50             & 0.005                  & 2000                  & 200                 & 0.03                           & hit@1              \\
WGCN                                          & TransE           & WN18RR           & 3              & 2               & 50             & 0.005                  & 5000                  & 200                 & 0.15                           & hit@1              \\
WGCN                                          & DistMult         & WN18RR           & 3              & 2               & 50             & 0.005                  & 5000                  & 200                 & 0.15                           & hit@1              \\
WGCN                                          & ConvE            & WN18RR           & 3              & 2               & 50             & 0.005                  & 5000                  & 200                 & 0.15                           & hit@1              \\ \bottomrule
\end{tabular}
}
\end{table*}

\end{document}